\documentclass{article}
\usepackage{spconf,amsmath,graphicx}

\usepackage{adjustbox} 
\usepackage{enumitem}
\setlist{nosep, leftmargin=14pt}

\usepackage{mwe} 

\usepackage{ctable} 
\usepackage{booktabs,tabularx}
\usepackage{multirow} 
\usepackage{array}
\usepackage{amsfonts}
\usepackage{dsfont}
\usepackage[htt]{hyphenat} 
\usepackage{subcaption}
\usepackage[colorlinks=true,
            linkcolor=magenta,
            urlcolor=gray,
            citecolor=cyan]{hyperref}
\usepackage{arydshln}
\NewDocumentCommand{\rot}{O{90} O{1em} m}{\makebox[#2][l]{\rotatebox{#1}{#3}}}%
\newcolumntype{?}{!{\vrule width 1pt}}

\title{A Relational-learning Perspective to Multi-label Chest X-ray Classification}
\name{Anjany Sekuboyina\sthanks{Work performed during an internship at NEC Labs.}$^{1, 2}$ \qquad Daniel O\~{n}oro-Rubio$^{1}$ \qquad Jens Kleesiek$^{3, 4}$ \qquad Brandon Malone$^{1}$}

\address{$^{1}$ NEC Laboratories Europe, Germany\\
         $^{2}$ Department of Informatics, Technical University of Munich, Germany \\
         $^{3}$ Institute for AI in Medicine (IKIM), University Hospital Essen, Germany \\
         $^{4}$ University of Duisburg-Essen, Germany}
\begin{document}
%
\maketitle
\begin{abstract}
Multi-label classification of chest X-ray images is frequently performed using discriminative approaches, i.e. learning to map an image directly to its binary labels. Such approaches make it challenging to incorporate auxiliary information such as annotation uncertainty or a dependency among the labels. Building towards this, we propose a novel knowledge graph reformulation of multi-label classification, which not only readily increases predictive performance of an encoder but also serves as a general framework for introducing new \emph{domain knowledge}. 

Specifically, we construct a multi-modal knowledge graph out of the chest X-ray images and its labels and pose multi-label classification as a \emph{link prediction} problem. Incorporating auxiliary information can then simply be achieved by adding additional nodes and relations among them. When tested on a publicly-available radiograph dataset (CheXpert), our relational-reformulation using a naive knowledge graph outperforms the state-of-art by achieving an area-under-ROC curve of 83.5\%, an improvement of $\sim 1\%$ over a purely discriminative approach.

\end{abstract}
\section{Introduction}
\label{sec:intro}
Chest radiography is one of the most common imaging modalities for diagnosing pathologies in lungs and heart, with multitudes of scans performed annually. Resulting high number of images require efficient screening. However, shortage of radiological experts, long working hours, manifestation of abnormalities with similar visual cues, etc. make accurate and efficient diagnosis difficult. Thus, there is a need for a computer-aided and explainable decision-support system in the clinical routine that can assist medical experts in reliably identifying pathological chest X-rays (CXR). This task is termed as \emph{multi-label CXR classification} (multiple co-occurring findings can be drawn from one radiograph).

Thanks to large-scale CXR datasets \cite{irvin2019chexpert,bustos2020padchest}, numerous deep learning-based approaches have been proposed to tackle the problem multi-label CXR classification. Two major factions of solutions proposed in the literature include: (1) \emph{ad-hoc} binary multi-label classification \cite{rajpurkar2017chexnet,chen2019dualchexnet} and (2) exploiting label-dependency-based classification \cite{chen2020label,pham2019interpreting,yao2017learning}. Approaches from the latter faction are interesting as they aim to exploit domain knowledge in addition to an image-to-label mapping. Yao \textit{et al.}~\cite{yao2017learning} propose a recurrent neural network-based decoder to learn the inter-dependency among labels. Pham \textit{et al.}~\cite{pham2019interpreting} construct a label tree and propose a hierarchical training regime of learning to classify the leaf nodes followed by the non-leaf ones, which is equivalent to conditional training. Chen \textit{et al.}~\cite{chen2020label}, on the other hand, propose to exploit label co-occurance. A graph based on the co-occurrence of the labels is used to enrich the feature representations learnt by a naive convolutional neural network (CNN) working on images.  

\begin{figure}
  \begin{minipage}[b]{1.0\linewidth}
  \centering
  \centerline{\includegraphics[width=7cm]{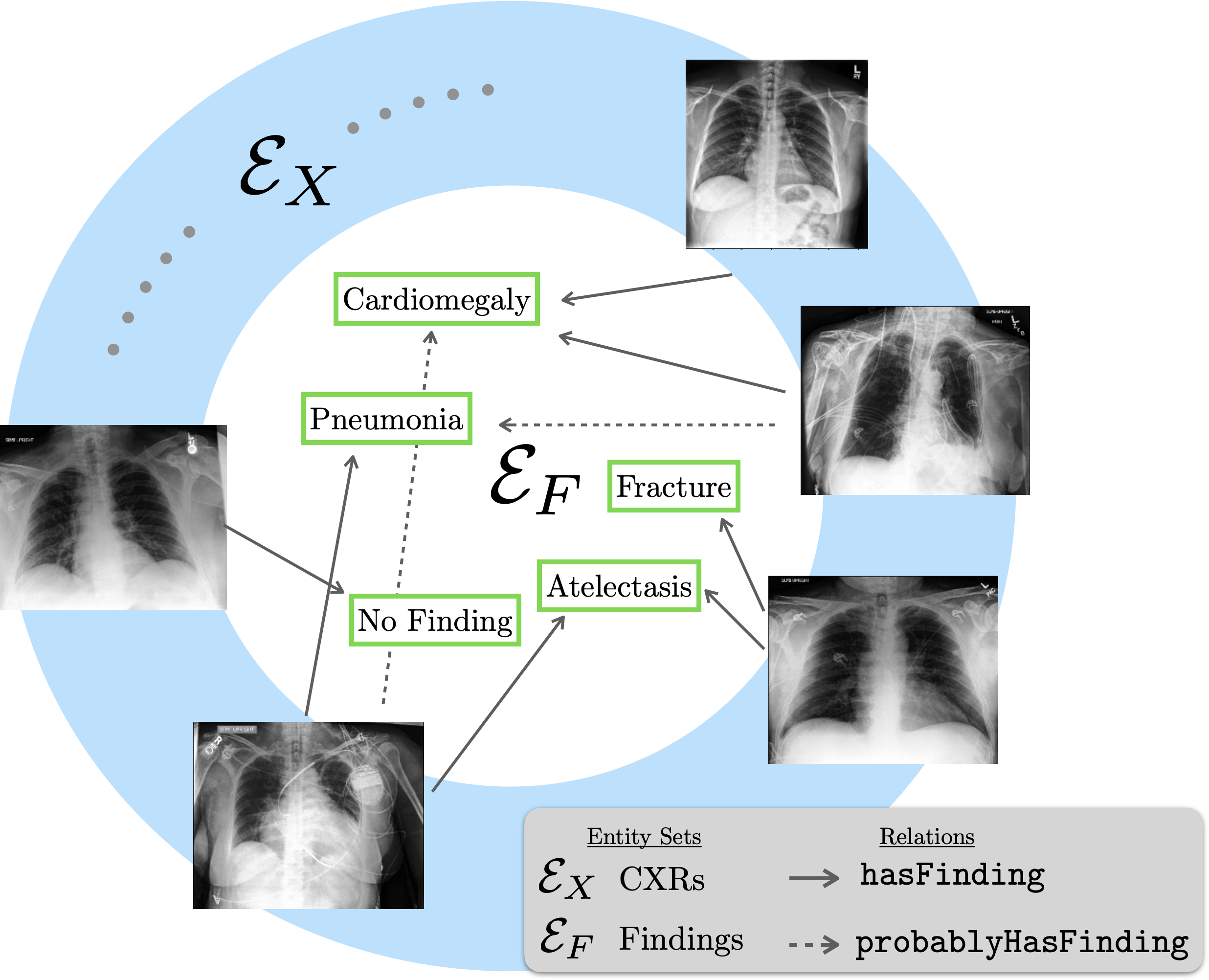}}
  \end{minipage}
  \caption{\small An overview of the \textbf{Radiological Knowledge Graph} constructed with the chest X-rays and `findings' as entities.}
  \label{fig:radkg}
\end{figure} 

Motivated by the idea of elegantly incorporating auxiliary information (e.g. domain knowledge) into the problem of multi-label CXR classification, we propose to reformulate the problem using a relational-learning perspective. Primarily, such a formulation involves a knowledge graph (KG) composed of the CXRs and the diseases (or findings) as the nodes (or entities) and the associations between them denoted by directed edges (or relations). We argue that this formulation is more flexible and show that it results in a better discriminator. New information can be added either as a new node or as a new relation.      

\subsection{Background: Link prediction in knowledge graphs}
In this section, we concisely introduce a knowledge graph (KG) and the link-prediction problem in the KG-realm. A knowledge graph $\mathcal{G}$ can formally be represented by a set of triplets $\{(s,r,o)\}$, where each triple denotes a `fact' in the real world. A triple captures the \emph{relation}, $r$, between a \emph{subject}, $s$ and an \emph{object}, $o$. The subject and object belong to a set of entities, $s,o\in\mathcal{E}$, and the relation belongs to a set of relations, $r\in\mathcal{R}$. For example, consider a sub-world with the following: $\mathcal{E}$ = \{\texttt{Kyoto}, \texttt{Japan}, \texttt{Murakami}\} and $\mathcal{R}$ = \{\texttt{bornIn}, \texttt{locatedIn}\}. A KG corresponding to this worlds could have triples such as (\texttt{Murakami}, \texttt{bornIn}, \texttt{Kyoto}), (\texttt{Murakami}, \texttt{bornIn}, \texttt{Japan}), (\texttt{Kyoto}, \texttt{locatedIn}, \texttt{Japan}), etc. 

Notice that such a KG can be incomplete due to reasons such as missing facts, corrupted data, etc., and it is of interest to \emph{complete} the graph. \emph{Link prediction} is one of the methods that address graph completion. It can be formalised as a triplet-ranking problem. It involves learning a scoring function, $\psi:\mathcal{E}\times\mathcal{R}\times\mathcal{E} \mapsto \mathbb{R}$, such that for a given triplet $x$, $\psi(x)$  takes a high value if $x$ is a true fact and a low value otherwise. 

\subsection{Contribution}

We propose to look at multi-label CXR classification as a link prediction problem in a radiological knowledge graph (RadKG). Example entity- and relation-sets for such a KG would be $\mathcal{E}$ = \{CXRs, diseases, other findings, patient reports, ...\} and $\mathcal{R}$ = \{\texttt{hasDisease}, \texttt{hasFinding}, \texttt{childOf}, ...\}. This formulation enables a seamless combination of multiple modalities of data as well as external domain knowledge such as disease ontologies \cite{bodenreider2004unified} and patient graphs \cite{friedman1990generalized}.    

Specifically, we combine the domains of relational machine learning with multi-label classification to makes three key contributions:
\begin{enumerate}
    \item We reformulate multi-label CXR classification as a link-prediction problem in a radiological knowledge graph.
    \item We show that our reformulation yields an area-under-ROC curve of 83.5\% compared to 83.2\% from prior state of the art \cite{chen2020label}, when tested on CheXpert \cite{irvin2019chexpert}. This is without incorporating any label dependency, unlike \cite{chen2020label}. 
    \item We demonstrate the flexibility of our KG representation by including additional information: (a) uncertain labels in the ground truth annotations and (b) the dependency among CXR findings. 
\end{enumerate}

  

\section{Method}
\label{sec:method}

We present our approach in two parts: First, we describe the construction of a simple radiological KG; second, we solve CXR multi-label classification using link-prediction in this KG.

\subsection{Radiological Knowledge Graph (RadKG)} 
The entity set $\mathcal{E}$ in RadKG is composed of the CXRs and the findings, $\mathcal{E} = \mathcal{E}_X + \mathcal{E}_F$. The CXR entity-subset consists of $m$ CXRs denoted by $X$, $\mathcal{E}_X = \{X_i\}_{i=1}^m$. The findings-subset is denoted by $\mathcal{E}_F = \{F_j\}_{j=1}^n$. In CheXpert \cite{irvin2019chexpert}, $F_j$ corresponds to findings such as `Pneumonia', `Edema', `Fracture' etc., with $n=14$. The image-to-label annotations in multi-label classification are denoted by relations between $\mathcal{E}_X$ and $\mathcal{E}_F$, $\mathcal{R} = \{\texttt{hasFinding}, \texttt{probablyHasFinding}\}$. Fig.~\ref{fig:radkg} illustrates RadKG. Specifically, \texttt{hasFinding} links a CXR and a finding when it is \emph{positively} annotated, while \texttt{probablyHasFinding} captures the uncertain annotations\footnote{CheXpert contains \emph{uncertain} annotations indicating \emph{findings} where the NLP-based automatic annotator was not confident about its prediction.}. Observe that lack of an edge between $X_i$ and $F_j$ implicitly implies a negative annotation.   

\begin{figure}
  \begin{minipage}[b]{1.0\linewidth}
  \centering
  \centerline{\includegraphics[width=8.5cm]{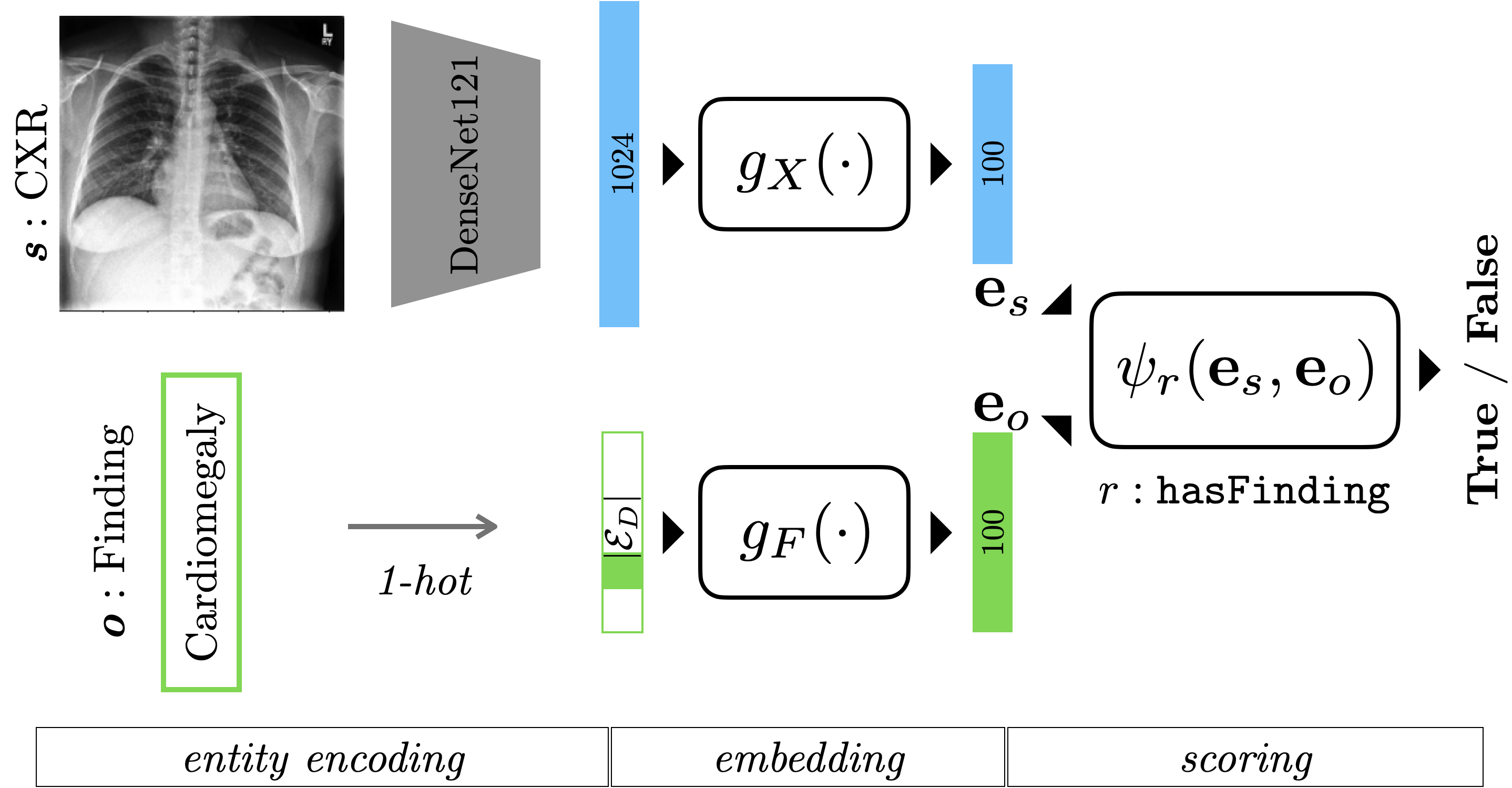}}
  \end{minipage}
  \caption{\small Schematic of the forward pass for scoring a given triple $(s,r,o)$ for the relation $r:$\texttt{hasFinding}.}
  \label{fig:link_pred}
\end{figure} 

\subsection{Link prediction on RadKG}
\noindent
\textbf{Formulating the problem.} 
A link-prediction model primarily operates on ascertaining the validity of triples. If a triple, $t = (s,r,o)$, exists in the KG, the model is supposed to predict a \texttt{TRUE} value and a \texttt{FALSE} otherwise. Fig.~\ref{fig:link_pred} illustrates this model for an example triple from the RadKG. The proposed link-prediction model consists of three modules: An \emph{encoding}, an \emph{embedding}, and a \emph{scoring} module. The encoding and embedding modules are responsible for mapping the entities in $t$ to an embedding space with representations denoted by $\mathbf{e}_s, \mathbf{e}_o \in \mathbb{R}^d$ ($d=100$, in this work). The scoring function, $\psi(s,r,o) = \psi_r(\mathbf{e}_s, \mathbf{e}_o) \in \mathbb{R}$, then scores $t$ at a high value if it is a \emph{positive} triple and vice versa. In the following sections, we detail the specifics of the three modules and describe the training and inference procedures.\\

\noindent
\textbf{Encoding the entities.} 
In a typical KG, the entities are abstract concepts. The \emph{encoding} module is responsible for representing these concepts as a mathematical entity. In RadKG, this representation concerns $\mathcal{E}_F$: a finding, $F_j$, is encoded into a one-hot vector, $\mathds{1}_j \in \mathbb{R}^{|\mathcal{E}_F|}$, active at the $j^\text{th}$ index. In this work, $|\mathcal{E}_F| = 14$ findings from CheXpert's label set. 


Additionally, RadKG also consists of CXRs ($\mathcal{E}_X$) which are visually-rich entities. Instead of one-hot representations as described above, we choose to encode $X_i$s using a convolutional neural network. Such domain-specific encoding not only avoids the \emph{ad-hoc}, orthogonal one-hot codes but enables efficient capture of the visual information in CXRs and results in similar CXRs having similar encodings. For this purpose, we extract the 1024-length feature representations from a 121-layered DenseNet (DenseNet121) pre-trained for multi-label CXR classification.\\

\noindent
\textbf{Embedding the entities.} 
The encoding module results in 1024-length codes for CXRs and $|\mathcal{E}_F|$-length codes for the findings. Note that these codes are static, especially since the DenseNet121 is pre-trained (offline) and the one-hot encoding for the findings has no learning component. Therefore, we incorporate an \emph{embedding} module which maps these codes to embedding spaces more suited for the link-prediction task. Denoting the encoded representations with $c_X$ and $c_F$, the embedding module consists of two function: $g_X(c_X): \mathbb{R}^{1024} \mapsto \mathbb{R}^d$ and $g_F(c_F): \mathbb{R}^{|\mathcal{E}_F|} \mapsto \mathbb{R}^d $. We denote the embedded representation by $\mathbf{e}$.\\

\noindent
\textbf{Scoring the triples.} 
Once the subject and object entities are embedded into the embedding space, recall that the triple $t$ needs to be scored for its validity. We investigate two scoring functions in this work, DistMult \cite{yang2014embedding} and ConvE \cite{dettmers2018conve}, formally denoted as follows:
\begin{align}
\text{DistMult:}~\psi_r(\mathbf{e}_s, \mathbf{e}_o)&=\langle \mathbf{e}_s, \mathbf{r}_r, \mathbf{e}_o \rangle,\\
\text{ConvE:}~\psi_r(\mathbf{e}_s, \mathbf{e}_o)&=f(\text{vec}(f([\overline{\mathbf{e}_s};\overline{\mathbf{r}_r}]*w))\mathbf{W})\mathbf{e}_o,
\end{align}
where $\mathbf{r}_r \in \mathbb{R}^d$ denotes the embedding for the relation $r$ and $\langle~\rangle$ denotes an inner product of the three vectors. The ConvE score function is relatively `convoluted', involving the reshaped and concatenated entity and relation embeddings, $\overline{\mathbf{e}_s}, \overline{\mathbf{r}_r} \in \mathbb{R}^{k_w, k_h}~(\text{s.t.~} k_w \times k_h = d; k_h = k_w = 10)$. The concatenated embeddings are passed through non-linear functions $f$ (=ReLU), convolved (kernel $w$ of size 5) and finally vectorised and passed through a fully connected-layer  with weight $\mathbf{W}$. In order to avoid clutter, we refer the reader to \cite{yang2014embedding} and \cite{dettmers2018conve} for the minutiae of the two scoring functions.\\

\noindent
\textbf{Learning to link.} 
In the proposed model, the entity and relation embeddings, along with the parameters of the scoring function parameters are to be learnt. They can be be trained with supervision once the scoring function is defined. This entails sampling valid and invalid samples from the training data and learning to assign high and low scores to them respectively. For this, we apply a sigmoid function on the raw scores, $p = \sigma(\psi_r(\mathbf{e}_s, \mathbf{e}_o))$, and minimise a binary-cross entropy loss, as formulated below:
$$
\mathcal{L}(p,y) = -y\log(p) - (1-y)\log(1-p), 
$$
where the target label, $y$, is $1$ if a triple is valid and $0$ otherwise. 

Once trained, we infer the labels of an unseen CXR, X$_{test}$, by scoring every completion from RadKG of the form ($X_{test}$, \texttt{hasFinding}, $F$?). The query-triple is scored for every $F_j \in \mathcal{E}_F$ as the object and positively annotated for those findings for which $\sigma(\psi_r(\cdot)) > \tau$, where $r$ is \texttt{hasFinding} and $\tau$ is chosen to appropriately tradeoff sensitivity and specificity.\\

\noindent
\textbf{Data.} CheXpert \cite{irvin2019chexpert} is a publicly-available chest radiograph dataset consisting of 224,315 lateral and frontal chest radiographs annotated for 14 findings with their \emph{presence}, \emph{absence}, or \emph{uncertain} presence. The data is split into three-folds of \emph{train}, \emph{validation}, and a \emph{test} sets with a $70:10:20$ proportion\footnote{Official CheXpert test set is not publicly available} stratified at patient-level. We work with the frontal CXRs padded to squares, resized to $320\times320$ pixels, and z-score normalised using the train-set's pixel mean and variance. Note that the test and validation sets do not contain CXRs with uncertain labels.\\

\section{Results}
\label{sec:exp}
\noindent
 The results of our experiments are tabulated in Table~\ref{tab:perf}. We use the area under ROC curve (AUC) as a metric to evaluate classification performance. All our experiments are evaluated over the three random data folds as described above and their mean is reported. We do not report standard deviation as it was less than 0.1\% in all cases. Similar to prior work, we perform two sets of experiments: first, we regard the uncertain labels as positives and second, as negatives. We compare our approach to two prior works \cite{irvin2019chexpert} and \cite{chen2020label}, the former using a typical CNN and the latter using a CNN whose feature are augmented with disease features learnt from a disease co-occurrence graph using a graph convolution network.\\
\emph{Baselines:} Recall that CXRs are encoded using a DenseNet121 trained for multi-label CXR classification. This naturally forms one of our baselines. Observe that the performance of our naive implementation is on par with the that of \cite{chen2020label}. Once the DenseNet is trained, the 1024-length features are stored as CXR codes and used for the consequent experiments. We then employ a 2-layered perceptron (MLP) with weight matrices of size $(1024\times100)$ and $(100\times14)$.\\    
\emph{Link prediction on RadKG:} Assessing the ability of a straightforward relational formulation, wherein we incorporate a KG with just the \texttt{hasFinding} relation, indicating positive annotations. The proposed approach consistently outperforms our baselines and prior work in both Tables~\ref{tab:perf}a and \ref{tab:perf}b, irrespective of the chosen scoring function. ConvE marginally outperforms DistMult, but its significance cannot be ascertained due to lack of sufficient samples. Interestingly, the trend of performances when Uncertain$\rightarrow$Positive being inferior to Uncertain$\rightarrow$Negative in the discriminative approaches is flipped in the relational approaches (0.835 vs. 0.833).\\
\emph{Incorporating domain knowledge:} We incorporate new knowledge into RadKG by adding two relations (cf. Table~\ref{tab:domain}): (1) \texttt{probablyHasFinding}, linking a CXR to a finding in cases where the ground truth has an uncertain annotation. (2) \texttt{coOccurs}, linking one finding to another based on their co-occurring probability. Directional co-occurrence is computed as in \cite{chen2020label} and one finding is said to `co-occur' with another if its probability of occurrence conditioned on the other is greater than 0.2. However, we do not observe an significant improvement due to this incorporation of domain knowledge. This behaviour is surprising and we attempt to explain it in the following section.\\

\begin{table}[t!]
\setlength{\tabcolsep}{0.8em}
\scriptsize
\renewcommand{\arraystretch}{1.5}
\caption{\small Performance of the proposed approach on a naive RadKG with only \texttt{hasFinding} relation. Uncertain labels in CheXpert ground truth are considered as positive annotations in (a) and as negative annotations in (b)}\label{tab:perf}
    \begin{subtable}{0.48\linewidth}\centering
    {
        \begin{tabular}{ c | c }
        \specialrule{.1em}{0em}{-.1em}
        Method & AUC \\ [0.25ex]
        \specialrule{.05em}{-0.1em}{0em}
        U\_Ones \cite{irvin2019chexpert} & 0.815 \\
        CheXGCN\_1s \cite{chen2020label} & 0.827 \\ 
        \hdashline
        DenseNet121 & 0.826 \\
        MLP &  0.827 \\
        RadKG+DistMult &  0.834 \\
        \textbf{RadKG+ConvE}  &  \textbf{0.835} \\[0.25ex]
        \specialrule{.1em}{0em}{0em}
        \end{tabular}
    }
    \caption{Uncertain $\rightarrow$ Positive}
    \end{subtable}%
    ~
    \begin{subtable}{0.48\linewidth}\centering
    {
         \begin{tabular}{ c | c }
        \specialrule{.1em}{0em}{-.1em}
        Method & AUC \\ [0.25ex]
        \specialrule{.05em}{-0.1em}{0em}
        U\_Zeros \cite{irvin2019chexpert} & 0.823 \\
        CheXGCN\_0s \cite{chen2020label} & 0.832 \\ 
        \hdashline
        DenseNet121 & 0.830 \\
        MLP & 0.831 \\
        RadKG+DistMult &  0.832 \\
        \textbf{RadKG+ConvE}  &  \textbf{0.833} \\[0.25ex]
        \specialrule{.1em}{0em}{0em}
        \end{tabular}
    }
    \caption{Uncertain $\rightarrow$ Negative}
    \end{subtable}

\end{table}

\begin{table}[t!]
\setlength{\tabcolsep}{0.8em}
\scriptsize
\centering
\renewcommand{\arraystretch}{1.5}

\caption{\small Performance of the proposed approach when RadKG is extended with extrinsic information in the form or two relations: \texttt{probablyHasDisease} and \texttt{coOccur}. Since the margins are insignificant, we report the standard deviations for a clearer picture.}\label{tab:domain}

\begin{tabular}{ c | c c }
    \specialrule{.1em}{0em}{-.1em}
    RadKG+ConvE &  Uncertain $\rightarrow$ $+$ve & Uncertain $\rightarrow$ $-$ve \\ [0.25ex]
    \specialrule{.05em}{-0.1em}{0em}
    + \texttt{probablyHasFinding} & 0.8339$\pm$0.0004 & 0.8330$\pm$0.0013 \\
     + \texttt{coOccur} & \textbf{0.8343}$\pm$0.0000 & 0.8332$\pm$0.0009 \\[0.25ex]
    \specialrule{.1em}{0em}{0em}
\end{tabular}

\end{table}

\section{Discussion} 
We discuss our approach in three parts: the DenseNet encoding, the relation formulation, and the domain-knowledge incorporation. The features extracted from the DenseNet can be classified into multiple classes with an AUC of 0.826 and 0.83 depending on the uncertain labels' mapping. Observe that a parametrically stronger classifier results in a marginal increase in performance of 0.001. On the other hand, the relational formulation using DistMult outperforms both these approaches (0.834 vs 0.826). Note that our DistMult formulation with $d=100$ has three embedding layers of size (1024$\times$100) for  images, (14$\times$100) for findings and (1$\times$100) for the \texttt{definitelyHas} relation, resulting in as many parameters as the MLP formulation. Moreover, the relational formulation appears to extract information lost during when the uncertain labels are mapped to positives, as evidenced by the boost in performance observed in this regard (cf.~Table~\ref{tab:perf}a). ConvE, on the other hand, with more parameters and an involved entity--relation interaction outperforms DistMult, as shown in \cite{dettmers2018conve}.

Incorporating two additional relations, \texttt{probablyHasFinding} and \texttt{coOccursWith}, did not yield an improvement. Non-success of \texttt{coOccursWith} can be attributed to the relatively fewer finding-to-finding triples compared to an image-to-finding triples, i.e $F_i$ has a far higher degree of edges linked to CXRs than to other $F_j$s. However, it is surprising that the effect of \texttt{probablyHasDisease} is minor. We assume this is due to the uncertain annotations in CheXpert being noise rather than useful information. Observe, in Table 3 of \cite{irvin2019chexpert}, that the experiment considering the uncertain labels as a separate class significantly outperforms others in only one out of the five diseases evaluated on. However, we expect that adding more reliable information, e.g. medical knowledge, will boost the performance further. \\

\noindent
\emph{Future work.} Proposed formulation of classification as a KG-completion problem opens up possibilities to fuse multi-modal data such as text reports and patient-population graphs. However, a encoding component pre-trained towards a certain task (multi-label classification, in our case) could result in representations that do not generalise well across tasks. Therefore, it is of interest to adopt a task-agnostic representation learning framework. Combining the encoding and embedding modules resulting in a fully-end-to-end formulation is also a future research direction.    

\section{Conclusion}
\label{sec:conc}
In this work, we present a relation-learning based reformulation of the multi-label CXR classification. A knowledge graph constructed with CXRs and `findings' as entities forms the core of our approach. We then pose the classification as a link-prediction problem in this KG. We demonstrate a superior performance of the proposed approach on publicly available CXR dataset, achieving an AUC of 83.5\%, outperforming state-of-art methods.\\

\section{Acknowledgements.}
The authors do not wish to declare any conflicts of interest. 

\section{Compliance With Ethical Standards.} 
This study was conducted retrospectively using human subject data made available in open-access as CheXpert \cite{irvin2019chexpert}. Explicit ethical approval from our end was not instructed by the data source.

\bibliographystyle{IEEEbib}
\bibliography{refs}

\begin{thebibliography}{10}

\bibitem{irvin2019chexpert}
Jeremy Irvin et~al.,
\newblock ``{CheXpert}: A large chest radiograph dataset with uncertainty
  labels and expert comparison,''
\newblock in {\em Proc. AAAI}, 2019.

\bibitem{bustos2020padchest}
Aurelia Bustos et~al.,
\newblock ``{PadChest}: A large chest x-ray image dataset with multi-label
  annotated reports,''
\newblock {\em Med Image Anal}, 2020.

\bibitem{rajpurkar2017chexnet}
Pranav Rajpurkar et~al.,
\newblock ``{CheXNet}: Radiologist-level pneumonia detection on chest {X-rays}
  with deep learning,''
\newblock {\em arXiv:1711.05225}, 2017.

\bibitem{chen2019dualchexnet}
Bingzhi Chen et~al.,
\newblock ``{DualCheXNet}: dual asymmetric feature learning for thoracic
  disease classification in chest {X-rays},''
\newblock {\em Biomed Signal Process Control}, 2019.

\bibitem{chen2020label}
Bingzhi Chen et~al.,
\newblock ``Label co-occurrence learning with graph convolutional networks for
  multi-label chest {X-ray} image classification,''
\newblock {\em IEEE Biomed Health Inform}, 2020.

\bibitem{pham2019interpreting}
Hieu~H Pham et~al.,
\newblock ``Interpreting chest {X-rays} via {CNNs} that exploit disease
  dependencies and uncertainty labels,''
\newblock 2019.

\bibitem{yao2017learning}
Li~Yao et~al.,
\newblock ``Learning to diagnose from scratch by exploiting dependencies among
  labels,''
\newblock {\em arXiv:1710.10501}, 2017.

\bibitem{bodenreider2004unified}
Olivier Bodenreider,
\newblock ``The unified medical language system ({UMLS}): integrating
  biomedical terminology,''
\newblock {\em Nucleic Acids Research}, 2004.

\bibitem{friedman1990generalized}
Carol Friedman et~al.,
\newblock ``A generalized relational schema for an integrated clinical patient
  database,''
\newblock in {\em Proc. Annu. Symp. Computer Appl. in Medical Care}, 1990.

\bibitem{yang2014embedding}
Bishan Yang et~al.,
\newblock ``Embedding entities and relations for learning and inference in
  knowledge bases,''
\newblock in {\em Proc. ICLR}, 2015.

\bibitem{dettmers2018conve}
Tim Dettmers et~al.,
\newblock ``Convolutional 2d knowledge graph embeddings,''
\newblock in {\em Proc. AAAI}, 2018.

\end{thebibliography}

\end{document}